\documentclass[preprint,12pt,authoryear]{elsarticle}

\usepackage{lineno}

\setcitestyle{round}

\usepackage{bm}
\usepackage{xurl}
\usepackage{amsmath,amssymb}
\usepackage{amsthm}
\usepackage{mathtools}
\usepackage{booktabs}
\usepackage[subrefformat=parens,skip=3pt]{subcaption}
\usepackage{tikz}
\usetikzlibrary{fit,arrows,calc,positioning}

\usepackage[hypertexnames=false]{hyperref}
\journal{Neural Networks}

\let\today\relax
\makeatletter
\def\ps@pprintTitle{%
    \let\@oddhead\@empty
    \let\@evenhead\@empty
    \def\@oddfoot{\footnotesize
         {
            \vbox{%
            \hbox{\raisebox{0.4ex}{\scalebox{0.7}{\textcopyright}} 2023. This manuscript version is made available under the CC-BY-NC-ND 4.0}%
            \hbox{\newline license \url{http://creativecommons.org/licenses/by-nc-nd/4.0/}}%
            }
        } \hfill\today}%
    \let\@evenfoot\@oddfoot
    }
\makeatother

\begin{document}

\begin{frontmatter}


\title{Stable Invariant Models via Koopman Spectra}

\author[1,2]{Takuya Konishi\corref{cor1}}
\ead{konishi@ist.osaka-u.ac.jp}
\cortext[cor1]{Corresponding author}

\author[1,2]{Yoshinobu Kawahara}

\affiliation[1]{
            organization={Graduate School of Information Science and Technology, Osaka University},
            addressline={1-5 Yamadaoka, Suita},
            state={Osaka},
            country={Japan}}

\affiliation[2]{
            organization={Center for Advanced Intelligence Project, RIKEN},
            addressline={1-4-1 Nihonbashi},
            city={Chuo-ku},
            state={Tokyo},
            country={Japan}}

\begin{abstract}
Weight-tied models have attracted attention in the modern development of neural networks. The deep equilibrium model (DEQ) represents infinitely deep neural networks with weight-tying, and recent studies have shown the potential of this type of approach. DEQs are needed to iteratively solve root-finding problems in training and are built on the assumption that the underlying dynamics determined by the models converge to a fixed \emph{point}. In this paper, we present the \emph{stable invariant model (SIM)}, a new class of deep models that in principle approximates DEQs under stability and extends the dynamics to more general ones converging to an invariant \emph{set} (not restricted in a fixed point). The key ingredient in deriving SIMs is a representation of the dynamics with the spectra of the Koopman and Perron--Frobenius operators. This perspective approximately reveals stable dynamics with DEQs and then derives two variants of SIMs. We also propose an implementation of SIMs that can be learned in the same way as feedforward models. We illustrate the empirical performance of SIMs with experiments and demonstrate that SIMs achieve comparative or superior performance against DEQs in several learning tasks.
\end{abstract}



\begin{keyword}
Neural networks \sep Deep learning \sep Dynamical systems \sep Spectral analysis



\end{keyword}

\end{frontmatter}


\section{Introduction}
\label{sec:intro}

A feedforward neural network learns a representation by explicitly iterating a number of layer-by-layer computations. Each layer performs a transformation of outputs from the previous layer, which is typically characterized by different sets of parameters among the layers. However, several recent studies have shown that models with weight-tying, i.e.\ the ones employing the same transformation in each layer, achieve results competitive with state-of-the-art performances \citep{BKK19b,DF19,DGV+19}. Motivated from this fact, \citet{BKK19a} recently proposed the deep equilibrium model (DEQ), which is equal to running an \emph{infinitely} deep feedforward model with weight-tying instead of using a finite number of layers. The models compute a representation by finding a fixed point (or an equilibrium point) with root-finding in practice, and are thus regarded as an instance of the so-called implicit-depth models such as neural ordinary differential equations~\citep{CRBD18}. The following studies on DEQs have shown capability in this type of approach in several learning tasks \citep{BKK20,WK20}.

The forward pass of DEQs and their variants involves solving root-finding problems, which can lead to a high computational time and tends to be unstable (regarding, for example, the sensitivity in hyper-parameter tuning and initialization). Hence, DEQs sometimes require extensive and time-consuming tuning to achieve strong performance and convergence to solutions \citep{LAG+20}. Additionally, DEQs assume that the underlying dynamics determined by the models converge to a fixed \emph{point}. However, as is often reported in papers on sequential neural models such as recurrent neural networks and also known in the scientific studies of brain activity, a broader class of convergence such as nonlinear oscillations could convey preferred capabilities in learning \citep{SM85,TIW+00,CCHC19,KZS20}.

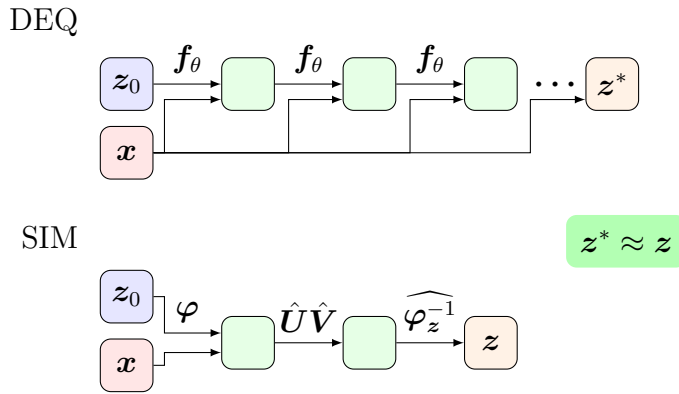
\begin{figure}[t]
    \centering
    \tikzstyle{b} = [rectangle, rounded corners, minimum width=0.55cm, minimum height=0.45cm,text centered, draw=black, fill=white]
    \tikzstyle{l} = [draw, -latex]
    \begin{tikzpicture}[auto]
        \node [b, minimum size=0.7cm, fill=red!10] (x_deq) {$\bm{x}$};
        \node [b, minimum size=0.7cm, fill=blue!10, above=0.2cm of x_deq] (z0_deq) {$\bm{z}_0$};
        \node [b, minimum size=0.7cm, fill=green!10, above=0.2cm of x_deq, right=0.9cm of z0_deq] (z1_deq) {};
        \node [b, minimum size=0.7cm, fill=green!10, above=0.2cm of x_deq, right=0.9cm of z1_deq] (z2_deq) {};
        \node [b, minimum size=0.7cm, fill=green!10, above=0.2cm of x_deq, right=0.9cm of z2_deq] (z3_deq) {};
        \node [right=0.07cm of z3_deq] (cdot) {$\bm{\cdots}$};
        \node [b, minimum size=0.7cm, fill=orange!10, above=0.2cm of x_deq, right=0.9cm of z3_deq] (z*_deq) {$\bm{z}^*$};
        \path [l] (z0_deq) -- node[above]{$\bm{f}_{\theta}$} (z1_deq);
        \path [l] (z1_deq) -- node[above]{$\bm{f}_{\theta}$} (z2_deq);
        \path [l] (z2_deq) -- node[above]{$\bm{f}_{\theta}$} (z3_deq);
        \path [l] (x_deq) -| ($(x_deq)!.4!(z1_deq.200)$) |- (z1_deq.210);
        \path [l] (x_deq) -| ($(x_deq)!.75!(z2_deq.200)$) |- (z2_deq.210);
        \path [l] (x_deq) -| ($(x_deq)!.84!(z3_deq.200)$) |- (z3_deq.210);
        \path [l] (x_deq) -| ($(x_deq)!.88!(z*_deq.200)$) |- (z*_deq.210);
        \node [above left=0.2cm of z0_deq] (DEQ) {DEQ};
        \node [b, minimum size=0.7cm, fill=blue!10, below=1.2cm of x_deq] (z0_sim) {$\bm{z}_0$};
        \node [b, minimum size=0.7cm, fill=red!10, below=0.2cm of z0_sim] (x_sim) {$\bm{x}$};
        \node [b, minimum size=0.7cm, fill=green!10, below right=-0.1cm and 0.9cm of z0_sim] (l1) {};
        \node [b, minimum size=0.7cm, fill=green!10, right=0.9cm of l1] (l2) {};
        \node [b, minimum size=0.7cm, fill=orange!10, right=0.9cm of l2] (z*_sim) {$\bm{z}$};
        \path [l] (x_sim) -| ($(x_sim)!.4!(l1.200)$) |- (l1.200);
        \path [l] (z0_sim) -| ($(z0_sim)!.4!(l1.140)$) |- node[above right]{$\bm{\varphi}$} (l1.160);
        \path [l] (l1) -- node[above]{$\hat{\bm{U}}\hat{\bm{V}}$} (l2);
        \path [l] (l2) -- node[above]{$\widehat{\bm{\varphi}_{\bm{z}}^{-1}}$} (z*_sim);
        \node [above left=0.2cm of z0_sim] (SIM) {SIM};
        \node [rectangle, minimum width=1.6cm, minimum height=0.7cm, rounded corners, text centered, fill=green!30, above right=0.9cm of z*_sim] (SIM) {$\bm{z}^* \approx \bm{z}$};
    \end{tikzpicture}
    \caption{Comparison of DEQs and single-tier SIMs. The notation is introduced in Sections~\ref{sec:background} and~\ref{sec:model}.}
    \label{illus:models}
\end{figure}

In this study, we propose a novel class of deep models, referred to as the \emph{stable invariant model (SIM)}. The key insight behind SIMs is to interpret the underlying dynamics regarding DEQs through the Koopman operator. The Koopman operator is a linear operator over functions defined on latent states of dynamics \citep{Koo31,Mez05}. Because of its linearity, we can capture inherent temporal and spatial patterns of dynamics by the representation of the spectra, i.e.\ eigenvalues and eigenvectors, of the corresponding Koopman operator. We first show that the spectra of the Koopman operator clarify stable dynamics with DEQs and then develop two variants of SIMs. The first models, single-tier SIMs, principally approximate DEQs under the stability. The resulting models, somewhat surprisingly, consist of only three-step transformations, although they approximate DEQs that are infinitely deep (see Figure~\ref{illus:models}). Moreover, the second models, two-tier SIMs, extend the dynamics to broader ones converging to an invariant \emph{set} (e.g.\ a set of points, curve, and more general manifold) by employing the connection between the Koopman and Perron--Frobenius operators. We further provide a practical scheme to implement SIMs so that they can be learned in the same manner as feedforward models. Finally, we illustrate the behaviors of SIMs with numerical experiments in supervised learning tasks. We demonstrate that our models achieve competitive or superior performances compared to DEQs with less computational time.

The remainder of this paper is organized as follows. First, in Section~\ref{sec:background}, we briefly review DEQs, and the Koopman and Perron-Frobenius operators. In Section~\ref{sec:model}, we propose SIMs along with the description of their resulting architectures and characteristics. We describe the related works in Section~\ref{sec:relwork} and investigate the empirical performance of our models in three learning tasks in Section~\ref{sec:expr}. We conclude this paper in Section~\ref{sec:conc}. The details of some equation derivations and experiments are presented in~\ref{sec:app-deriv-limit} and~\ref{sec:app-ex}, respectively.

\section{Background}
\label{sec:background}

\subsection{Deep Equilibrium Models}
\label{ssec:equilibrium}

One of the core ideas in DEQs is weight-tying, i.e.\ the same set of parameters is shared across the layers of a deep network. Formally, DEQs consider an $L$-layer weight-tied transformation with shared parameters $\theta$:
\begin{align}
\label{eq:deq}
\bm{z}_{l+1} = \bm{f}_\theta(\bm{z}_l,\bm{x}),~~~l=0,1,\ldots,L-1,
\end{align}
where $\bm{x} \in \mathbb{R}^D$ is the input to the model, $\bm{z}_l \in \mathbb{R}^d$ is the hidden state of the $l$-th layer, and $\bm{f}_\theta \colon \mathbb{R}^{d+D} \to \mathbb{R}^{d}$ is a continuous function. DEQs suppose that stacking such layers infinitely tends to a fixed point:
\begin{align}
\label{eq:deq-limit}
\lim_{l\to\infty}\bm{z}_l = \lim_{l\to\infty}\bm{f}_\theta^l(\bm{z}_0,\bm{x}) \coloneqq \bm{f}_\theta(\bm{z}^*,\bm{x}) = \bm{z}^*.
\end{align}
The forward pass of DEQs uses root-finding algorithms to directly compute the fixed point $\bm{z}^*$ by solving the equation $\bm{z}^*=\bm{f}_\theta(\bm{z}^*,\bm{x})$. Then, $\bm{z}^*$ is transformed to an output $\bm{y}$ by a function $\bm{h}$ as $\bm{y} = \bm{h}(\bm{z}^*)$. We can train DEQs with backpropagation by computing the gradient of the fixed point through implicit differentiation \citep{KP03}.

A series of transformations in Eq.~\eqref{eq:deq} can be viewed as a discrete-time nonlinear dynamical system where the hidden state $\bm{z}_l$ is a state vector at step $l$. The underlying dynamics are determined by the transformation $\bm{f}_\theta$, which is affected by the input $\bm{x}$ at every step.

\subsection{Koopman and Perron--Frobenius Operators}
\label{ssec:koopman}

We briefly overview the Koopman and Perron--Frobenius operators. Please see other references, e.g.\ \citep{MMS20}, for more information.

\paragraph{Definitions}
Consider a discrete-time nonlinear dynamical system:\@ $\bm{z}_{t+1}$$\,=\bm{f}(\bm{z}_t)$, defined on a state space~$\mathbb{S}$$\,\subset$$\,\mathbb{R}^d$, where $\bm{z}_t$$\, \in \mathbb{S}$ is the state vector at time $t$, and $\bm{f} \colon \mathbb{S} \to \mathbb{S}$ is a (possibly, nonlinear) state-transition function. Let $g \in \mathcal{G}$ be an \emph{observable}, which is a scalar complex-valued function on $\mathbb{S}$ in some (Banach) space $\mathcal{G}$.  The \emph{Koopman operator} $\mathcal{K} \colon \mathcal{G} \to \mathcal{G}$ is defined through the following composition:
\begin{align*}
\left(\mathcal{K}g\right)(\bm{z}) = g\left(\bm{f}(\bm{z})\right),
\end{align*}
where $\bm{z} \in \mathbb{S}$ is a state vector. $\mathcal{K}$ acts on observables and maps $g$ to a new function $\mathcal{K}g$. Although the dynamics described by $\bm{f}$ may be nonlinear, $\mathcal{K}$ is linear and infinite-dimensional.

The Perron--Frobenius operator is often used to describe the transition of the density over the state of dynamical systems~\citep{LM94, Gas98, CAM+20}. Given a measure space $(\mathbb{S}, \mathbb{A}, \mu)$ that consists of a state space $\mathbb{S}$, $\sigma$-algebra $\mathbb{A}$, and measure $\mu$, we suppose that $\bm{f}$ is nonsingular if $\mu(\bm{f}^{-1}(A)) = 0$ for a Borel set $A \in \mathbb{A}$ such that $\mu(A)=0$, where $\bm{f}^{-1}(A)$ is the preimage of $\bm{f}$ given $A$. Let $p \in \mathcal{L}^1$ denote a density function on $\mathbb{S}$ in the space of absolutely integrable functions $\mathcal{L}^1$. The \emph{Perron--Frobenius operator} $\mathcal{P} \colon \mathcal{L}^1 \to \mathcal{L}^1$ acts on densities and is defined as
\begin{align*}
\int_A (\mathcal{P} p)(\bm{z}) d\mu  = \int_{\bm{f}^{-1}(A)} p(\bm{z}) d\mu, \;\;\;\; A \in \mathbb{A}.
\end{align*}
It should be noted that the Koopman (or Perron--Frobenius) operator is the adjoint of the Perron--Frobenius (or Koopman) operator for appropriately defined spaces.

\paragraph{Koopman Spectrum}
Because the Koopman operator $\mathcal{K}$ (and also Perron--Frobenius operator) is linear, it can be characterized by spectral properties. We assume $\mathcal{K}$ has only point spectra and also has non-trivial eigenfunctions. Let $\lambda_j \in \mathbb{C}$, $j = 1,2,\ldots$, be the eigenvalue of $\mathcal{K}$. The eigenfunction $\phi_j \colon \mathbb{S} \to \mathbb{C}$ for $\lambda_j$ satisfies the relation
\begin{align*}
\left(\mathcal{K}\phi_j\right)(\bm{z}) = \phi_j \left( \bm{f}(\bm{z}) \right) = \lambda_j \phi_j(\bm{z}).
\end{align*}
$\lambda_j$ and $\phi_j$ are called the Koopman eigenvalue and Koopman eigenfunction, respectively. If an observable $g$ is in the subspace of $\mathcal{G}$ spanned by the Koopman eigenfunctions $\left\{\phi_j\right\} \coloneqq \left\{\phi_j \mid j=1,2,\ldots \right\}$, the observable can be represented as
\begin{align*}
\left(\mathcal{K}g\right)(\bm{z}) = g \left( \bm{f} (\bm{z}) \right) = \sum_{j=1}^{\infty} \lambda_j v_j \phi_j(\bm{z}),
\end{align*}
where the coefficient $v_j \in \mathbb{C}$, $j=1,2,\ldots$, is referred to as the Koopman mode associated with $g$. The subspace spanned by $\left\{\phi_j\right\}$ is invariant under the Koopman operator, i.e., the observables in the subspace remain in the subspace after being acted by $\mathcal{K}$.

\paragraph{Finite-Dimensional Approximation}
Consider a subspace of observables which is spanned by $N$ basis functions $\left\{ \varphi_j \right\} \coloneqq \left\{\varphi_j \colon \mathbb{S} \to \mathbb{C} \mid j=1,2,\ldots,N\right\}$. If $g$ exists on the subspace, then $g$ is represented as a linear combination of the basis functions, i.e.\ $g(\bm{z}) = \bm{w}^{\top}\bm{\varphi}(\bm{z})$, where we denote the concatenation of the basis functions as a vector-valued one $\bm{\varphi} = (\varphi_1,\dots,\varphi_N)^{\top} \colon \mathbb{S} \to \mathbb{C}^N$, and $\bm{w} \in \mathbb{C}^N$ is the coordinate of $g$ on the subspace. By projecting the action of $\mathcal{K}$ onto the span of $\left\{\varphi_j\right\}$, we approximate $\mathcal{K}$ with another linear operator $\mathcal{K}_N$. This approximates the Koopman operator $\mathcal{K}$ and satisfies

\begin{align*}
(\mathcal{K}_N g)(\bm{z}) = (\bm{Kw})^{\top} \bm{\varphi}(\bm{z}),
\end{align*}
where $\bm{K} \in \mathbb{C}^{N \times N}$ is referred to as the Koopman matrix. The above shows that a temporal evolution of $g$ with $\mathcal{K}_N$ is represented by applying $\bm{K}$ to the coordinate $\bm{w}$. Therefore, the Koopman matrix owns the one-to-one correspondence to $\mathcal{K}_N$. Moreover, the Koopman eigenvalues, eigenfunctions, and modes of $\mathcal{K}_N$ can be obtained from the eigenvalues, right-eigenvectors, and left-eigenvectors of $\bm{K}$, respectively.

Additionally, the Koopman matrix provides an approximation of the dynamics through the basis functions:
\begin{align}
\label{eq:lift}
\bm{\varphi}\left(\bm{f}(\bm{z})\right)
\approx (\bm{K}^{\top} \bm{\varphi})(\bm{z})
\coloneqq (\bm{A} \bm{\varphi})(\bm{z}),
\end{align}
where $\bm{A}$ is the transpose of the Koopman matrix, i.e.\ $\bm{A} =\bm{K}^{\top}$. Eq.~\eqref{eq:lift} indicates that a temporal evolution with $\bm{f}$ can be approximated by a temporal evolution with the finite and linear dynamics described by $\bm{A}$ over a \emph{lifted} space with $\bm{\varphi}$. The equation holds if $\mathcal{K} = \mathcal{K}_N$.

It should be noted that, if $g$ is a real-valued function, then it is sufficient to consider real-valued ones for the corresponding quantities (such as $\varphi_j$, $\bm{w}$ and $\bm{K}$).

\section{Stable Invariant Models}
\label{sec:model}

In this section, we introduce our proposed SIMs. The models are motivated by two problems concerning DEQs. First, DEQs work under the assumption that the underlying dynamics are stable, i.e.\ Eq.~\eqref{eq:deq-limit} holds for any input and initial state. However, it is in general hard to assess whether nonlinear dynamics satisfy the assumption. We address this problem by approximately specifying the stable dynamics using the Koopman spectrum, which leads our first single-tier SIMs by identifying the convergent behavior for a fixed point.

The second problem is that DEQs only consider convergence to a fixed point. In the literature on dynamical systems, one often considers more general convergence to an invariant set: for a dynamical system on a state space $\mathbb{S}$, a set $S \subset \mathbb{S}$ is said to be an invariant set if any trajectory starting in $S$ remain in $S$. The convergence to an invariant set allows for the dynamics that oscillate on the set. Notable examples include limit cycles, tori, and other nonlinear oscillations~\citep{Str15}. We also refer to dynamics as stable if the dynamics converge to a bounded invariant set for any input and initial state and construct our second two-tier SIMs by incorporating only the dynamics converging to an invariant set via the Koopman and Perron--Frobenius operators.

\subsection{Approximating DEQs}

We begin by considering the representations of DEQs with the Koopman operator. However, it should be noted that $\bm{f}_\theta$ in Eq.~\eqref{eq:deq} has another vector $\bm{x}$ as an input differently from $\bm{f}$ in Section~\ref{ssec:koopman}. Although there are several ways to define the Koopman operator for such a case, we consider the following Koopman operator $\mathcal{K}$ acting on the observable $g$ of both the hidden state $\bm{z}$ and input $\bm{x}$ \citep{PBK18}:
\begin{align}
\label{eq:koopman-deq}
(\mathcal{K}g)(\bm{z}, \bm{x}) = g(\bm{f}_\theta(\bm{z}, \bm{x}), \bm{x}).
\end{align}
The action of this operator is restricted so that $\bm{x}$ is maintained at the same point. Even if $\bm{x}$ is injected, the properties of the Koopman operator discussed in Section~\ref{ssec:koopman} still hold.

Hereafter, we consider a real-valued observable $g$. First, let $\varphi_j(\bm{z}, \bm{x}) \colon \mathbb{R}^{d+D}$ $\to \mathbb{R}$ be $N$ real-valued basis functions ($j=1,\ldots,N$). The concatenation is given by a vector-valued basis function $\bm{\varphi}$. If Eq.~\eqref{eq:deq} converges to a fixed point $\bm{z}^*$ as in Eq.~\eqref{eq:deq-limit}, we can approximate $\bm{z}^*$ with $\bm{x}$ by the Koopman operator of Eq.~\eqref{eq:koopman-deq} as
\begin{align}
\label{eq:deq-basis-approx}
(\bm{z}^*, \bm{x})
&=\left(\lim_{l\to\infty} \bm{f}_{\theta}(\bm{z}_l,\bm{x}), \bm{x} \right) \nonumber \\
&=\left(\lim_{l\to\infty} \bm{f}_{\theta}^l(\bm{z}_0,\bm{x}), \bm{x} \right) \nonumber \\
&\approx \widehat{\bm{\varphi}^{-1}} \left( \bm{\varphi} \left(\lim_{l\to\infty} \bm{f}_{\theta}^l(\bm{z}_0,\bm{x}), \bm{x} \right) \right) \nonumber\\
&\approx \widehat{\bm{\varphi}^{-1}} \left(\lim_{l\to\infty} \bm{A}^l \bm{\varphi}(\bm{z}_0,\bm{x}) \right).
\end{align}
Here, we define a function $\widehat{\bm{\varphi}^{-1}}$ in the first approximation. The equation holds if there exists the inverse function $\bm{\varphi}^{-1}$ and $\widehat{\bm{\varphi}^{-1}} = \bm{\varphi}^{-1}$. Strictly invertible $\bm{\varphi}$ with respect to all possible inputs and outputs is rather restrictive and expensive in practice. We relax the invertibility by supposing $\widehat{\bm{\varphi}^{-1}}$ as a surrogate function that approximately models the output--input relations in the subspace where most of the inputs and outputs are distributed. The second approximation of Eq.~\eqref{eq:deq-basis-approx} follows the finite-dimensional approximation of the Koopman operator through the lifted dynamics as described in Eq.~\eqref{eq:lift}.
The basis functions need higher expressiveness to better approximate the subspace of the observables that the Koopman operator acts. By dividing the surrogate function $\widehat{\bm{\varphi}^{-1}}$ into two parts corresponding to $\bm{z}$ and $\bm{x}$, i.e.\ $\widehat{\bm{\varphi}^{-1}} = (\widehat{\bm{\varphi}_{\bm{z}}^{-1}}, \widehat{\bm{\varphi}_{\bm{x}}^{-1}})$, the fixed point $\bm{z}^*$ can be approximated more directly as
\begin{align}
\label{eq:deq-approx}
\bm{z}^* &\approx \widehat{\bm{\varphi}_{\bm{z}}^{-1}} \left(\lim_{l\to\infty} \bm{A}^l \bm{\varphi}(\bm{z}_0,\bm{x}) \right).
\end{align}

\subsection{Convergent Behavior via Koopman Spectra}

The approximation~\eqref{eq:deq-approx} assumes that DEQs converge to a fixed point. However, whether a DEQ converges to a fixed point depends on the behavior of the underlying dynamics with the DEQ. The approximation \eqref{eq:deq-approx} allows us to characterize the convergent behavior of a DEQ via eigenvalues of the corresponding $\bm{A}$.

First, we denote by $\lambda_j \in \mathbb{C}$ for $j=1,2,\ldots,N$ the eigenvalues of $\bm{A}$. It should be noted that the eigenvalues can be complex values even though $\bm{A}$ is a real matrix (because it is not necessarily symmetric). For any real matrix $\bm{A}$, there exists a nonsingular matrix $\bm{U} \in \mathbb{R}^{N \times N}$ that consists of the generalized-eigenvectors including the ordinal eigenvectors of $\bm{A}$. We represent $\bm{U}$ and $\bm{U}^{-1}$ with $N$ vectors, respectively, as $\bm{U} = (\bm{u}_1,\ldots,\bm{u}_N)$ and $\bm{U}^{-1} = (\bm{v}_1,\ldots,\bm{v}_N)^{\top}$, where $\bm{u}_j$ is associated with a generalized-eigenvector of $\lambda_j$ if $\lambda_j$ is real, and the real or imaginary part of a generalized-eigenvector of $\lambda_j$ if $\lambda_j$ is non-real. Additionally, let $\rho(\bm{A})$ be the spectral radius of $\bm{A}$, i.e.\ $\rho(\bm{A}) \coloneqq \max\{|\lambda_1|,\ldots,|\lambda_N|\}$.
We can then classify the convergent behavior of the lifted dynamics into the following four cases:
\renewcommand{\theenumi}{(\roman{enumi})}
\begin{enumerate}

\item \label{enu:case-origin} If $\rho(\bm{A}) < 1$, then the dynamics converge to the origin:
\begin{align*}
\lim_{l\to\infty} \bm{A}^l \bm{\varphi}(\bm{z}_0,\bm{x}) = \bm{0}.
\end{align*}

\item \label{enu:case-point} If $\rho(\bm{A}) = 1$, all the eigenvalues with $|\lambda_j|=1$ take the values 1, and their corresponding eigenvectors are linearly independent, then the lifted dynamics converge to a fixed point. That is, if we denote by $J_1 = \{j \mid \lambda_j = 1\}$ the index set of such eigenvalues, then we have
\begin{align}
\label{eq:case-point}
\lim_{l\to\infty} \bm{A}^l \bm{\varphi}(\bm{z}_0,\bm{x})
= \sum_{j \in J_1} \bm{u}_j \bm{v}_j^{\top} \bm{\varphi}(\bm{z}_0,\bm{x}).
\end{align}

\item \label{enu:case-set} If $\rho(\bm{A}) = 1$ and the eigenvectors with eigenvalues of $|\lambda_j|=1$ are linearly independent, then the lifted dynamics do not converge to a point but oscillates in the state space. More concretely, if we denote $\lambda_j = \alpha_j + i \beta_j$, $J_2 = \{j \mid \lambda_j = -1 \}$, and $J_3 = \{ (j,k) \mid |\lambda_j| = |\lambda_k| = 1, \beta_j = \beta_k \neq 0, \lambda_k = \overline{\lambda_j} \}$, then we have
\begin{align}
\begin{split}
\lim_{l\to\infty} \bm{A}^l \bm{\varphi}(\bm{z}_0,\bm{x})
& = \lim_{l\to\infty} \Biggl( \sum_{j \in J_1} \bm{u}_j \bm{v}_j^{\top}
    + \sum_{j \in J_2} (-1)^l \bm{u}_j \bm{v}_j^{\top} \\
    &  + \sum_{(j,k) \in J_3} \big(\cos(l\Delta_j)\bm{u}_{j}-\sin(l\Delta_k)\bm{u}_{k}\big) \bm{v}_{j}^{\top} \\
    & + \big(\sin(l\Delta_j)\bm{u}_{j}+\cos(l\Delta_k)\bm{u}_{k}\big) \bm{v}_{k}^{\top} \Biggr)
    \bm{\varphi}(\bm{z}_0,\bm{x}),
\end{split}
\label{eq:case-set}
\end{align}
where $i$ is the imaginary unit, and $\Delta_j = \arctan(\beta_j/\alpha_j)$. $J_3$ is the set of index pairs whose eigenvalues are conjugated. It should be noted that there always exists a conjugate eigenvalue for every non-real eigenvalue when $\bm{A}$ is real.

\item \label{enu:case-unstable} Otherwise, at least one element among the states of the lifted dynamics diverges.
\end{enumerate}
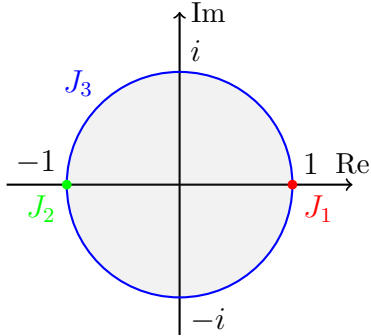
\begin{figure}[t]
    \centering
    \begin{tikzpicture}
        \begin{scope}[thick,font=\small][set layers]
            \draw [->] (-2.3,0) -- (2.3,0) node [above] {Re};
            \draw [->] (0,-2.0) -- (0,2.3) node [right] {Im};
        \end{scope}
        \node [above right, black] at (1.5,0) {$1$};
        \node [above left, black] at (-1.5,0) {$-1$};
        \node [above right, black] at (0,1.5) {$i$};
        \node [below right, black] at (0,-1.5) {$-i$};
        \path [draw=blue, thick, fill=gray, fill opacity=0.1] (0,0) circle (1.5);
        \node [blue] at (-1.35,1.35) {$J_3$};
        \path [draw=red, thick, fill=red] (1.5,0) circle (0.05);
        \node [below right, red] at (1.5,0) {$J_1$};
        \path [draw=green, thick, fill=green] (-1.5,0) circle (0.05);
        \node [below left, green] at (-1.5,0) {$J_2$};
    \end{tikzpicture}
    \caption{Areas where eigenvalues corresponding to $J_1$, $J_2$, and $J_3$ are located in the complex plane.}
    \label{illus:cplane}
\end{figure}
We describe the derivations of the four cases in~\ref{sec:app-deriv-limit}. Figure~\ref{illus:cplane} shows the complex plane and coordinates of the eigenvalues described in the above cases. The red and green dots, and the blue line denote the locations of the eigenvalues that correspond to $J_1$, $J_2$, and $J_3$, respectively. The gray area shows the area where the absolute values of the eigenvalues are less than 1. The first three cases approximately correspond to the stable dynamics with the DEQ. Case~\ref{enu:case-origin} converges to the origin regardless of $\bm{x}$, which is useless for any learning problems. In contrast, case~\ref{enu:case-point} converges to a fixed point that reflects $\bm{x}$. Case~\ref{enu:case-set} does not converge to any fixed point but oscillates on a manifold in the state space, where the terms corresponding to $J_2$ and $J_3$ respectively include coefficients such as $(-1)^l$ and $\cos(l\Delta_j)$ that neither diverge nor converge. Lastly, case~\ref{enu:case-unstable} diverges and the corresponding dynamics are not stable.

\subsection{Model Description}

We derive SIMs based on the above analysis of DEQs from the perspective of the Koopman spectrum. We first describe the variant that approximates DEQs by leveraging case~\ref{enu:case-point}. Moreover, we present another variant of SIMs that incorporates a broader class of dynamics by utilizing case~\ref{enu:case-set}.

\paragraph{Fixed Point}
Case~\ref{enu:case-point} represents the dynamics that converge to a fixed point as in Eq.~\eqref{eq:case-point}.
Because DEQs assume that the underlying dynamics converge towards a fixed point, which are covered by case~\ref{enu:case-point},
we can approximate the fixed point of a DEQ by plugging Eq.~\eqref{eq:case-point} into Eq.~\eqref{eq:deq-approx}:
\begin{align}
\label{eq:single-tier-model}
    \bm{z}^*
    \approx \widehat{\bm{\varphi}_{\bm{z}}^{-1}} \left(\sum_{j \in J_1} \bm{u}_j \bm{v}_j^{\top} \bm{\varphi}(\bm{z}_0,\bm{x}) \right)
    = \widehat{\bm{\varphi}_{\bm{z}}^{-1}} \left(\hat{\bm{U}}\hat{\bm{V}} \bm{\varphi}(\bm{z}_0,\bm{x}) \right),
\end{align}
where $\hat{\bm{U}} \in \mathbb{R}^{N \times K}$ and $\hat{\bm{V}} \in \mathbb{R}^{K \times N}$ consist of the vectors that are the rows of $\bm{U}$ and $\bm{V}$ whose indices belong to $J_1$ and $K$ is the size of $J_1$. Interestingly, this approximation implies that a fixed point of DEQs can be represented as a finite-depth model that consists of three-step transformations if the basis functions can approximate the subspace of the observables that the corresponding Koopman operator acts. We refer to the right-hand side of Eq.~\eqref{eq:single-tier-model} \emph{single-tier} SIMs. This comes from the fact that a state of the dynamics is lifted to another tier with $\bm{\varphi}$.

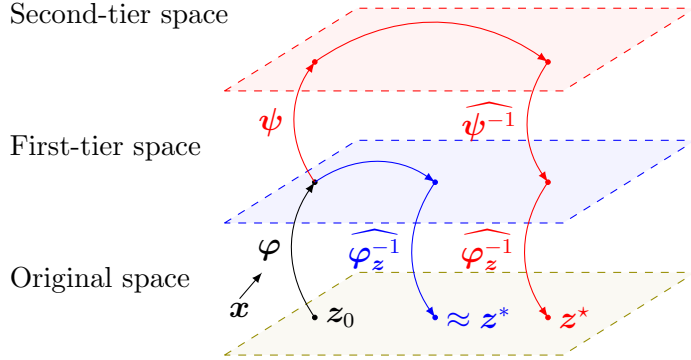
\begin{figure}[t]
    \centering
    \begin{tikzpicture}
        \begin{scope}[]
          \draw[olive, dashed, thin, fill=olive, fill opacity=0.05, style={yslant=0.0,xslant=1.6}] (0,0) rectangle (4.5,1.1);
          \node [right, black, align=left] at (-3.,1.) {\small Original space};
        \end{scope}
        \begin{scope}[yshift=50]
          \draw[blue, dashed, thin, fill=blue, fill opacity=0.05, style={yslant=0.0,xslant=1.6}] (0,0) rectangle (4.5,1.1);
          \node [right, black, align=left] at (-3.,1.) {\small First-tier space};
        \end{scope}
        \begin{scope}[yshift=100]
          \draw[red, dashed, thin, fill=red, fill opacity=0.05, style={yslant=0.0,xslant=1.6}] (0,0) rectangle (4.5,1.1);
          \node [right, black, align=left] at (-3.,1.) {\small Second-tier space};
        \end{scope}
        \path [draw=black, fill=black] (1.2,0.5) circle (0.03) node[right] {$\bm{z}_0$};
        \node [black] at (0.2,0.6) {$\bm{x}$};
        \draw[thin, -latex] (0.2,0.75) to (0.5,1.1);
        \draw[thin, -latex] (1.2,0.5) to [bend left=35] node[left] {$\bm{\varphi}$} (1.2,2.3);
        \path [draw=black, fill=black] (1.2,2.3) circle (0.03);
        \draw[red, thin, -latex] (1.2,2.3) to [bend left=35] node[left] {$\bm{\psi}$} (1.2,3.9);
        \path [draw=red, fill=red] (1.2,3.9) circle (0.03);
        \draw[blue, thin, -latex] (1.2,2.3) to [bend left=35] (2.8,2.3);
        \path [draw=blue, fill=blue] (2.8,2.3) circle (0.03);
        \draw[blue, thin, -latex] (2.8,2.3) to [bend right=35] node[left] {$\widehat{\bm{\varphi_z}^{-1}}$} (2.8,0.5);
        \path [draw=blue, fill=blue] (2.8,0.5) circle (0.03) node[blue, right] {$\approx \bm{z}^*$};
        \draw[red, thin, -latex] (1.2,3.9) to [bend left=35] (4.3,3.9);
        \path [draw=red, fill=red] (4.3,3.9) circle (0.03);
        \draw[red, thin, -latex] (4.3,3.9) to [bend right=35] node[left] {$\widehat{\bm{\psi}^{-1}}$} (4.3,2.3);
        \path [draw=red, fill=red] (4.3,2.3) circle (0.03);
        \draw[red, thin, -latex] (4.3,2.3) to [bend right=35] node[left] {$\widehat{\bm{\varphi_z}^{-1}}$} (4.3,0.5);
        \path [draw=red, fill=red] (4.3,0.5) circle (0.03) node[red, right] {$\bm{z}^{\star}$};
    \end{tikzpicture}
    \caption{Overview of SIMs.}
    \label{illus:overview}
\end{figure}

\paragraph{Invariant Set}
Case~\ref{enu:case-set} represents the dynamics that converge to an invariant set given by Eq.~\eqref{eq:case-set}. While such dynamics do not converge to any single fixed point, it is necessary to characterize case~\ref{enu:case-set} with some representative point to obtain a trainable model. We here focus on invariant sets being characterized by spectra with eigenvalue~1 of the corresponding Perron--Frobenius operator \citep{BS08,FP09}. This viewpoint suggests that the spectra can encode the information of convergent trajectories in case~\ref{enu:case-set}.

Now, let $\bm{\psi}$ be the (finite-dimensional) basis functions of an embedding of the state into some Hilbert space that encodes the density defining the corresponding Perron--Frobenius operator. Building upon the above insight, we propose to represent case~\ref{enu:case-set} as the following extended equilibrium over the states on one more lifted space similar to case~\ref{enu:case-point}:
\begin{align}
\label{eq:two-tier-model}
\begin{split}
\bm{z}^{\star}
&\coloneqq \widehat{\bm{\varphi}_{\bm{z}}^{-1}} \left( \widehat{\bm{\psi}^{-1}} \left(\sum_{j \in J_1^{\prime}} \bm{u}_j^{\prime} \bm{v}_j^{\prime\top} \bm{\psi} \left(\bm{\varphi}(\bm{z}_0,\bm{x}) \right) \right) \right) \\
&= \widehat{\bm{\varphi}_{\bm{z}}^{-1}} \left( \widehat{\bm{\psi}^{-1}} \left( \hat{\bm{U}}^{\prime}\hat{\bm{V}}^{\prime} \bm{\psi} \left(\bm{\varphi}(\bm{z}_0,\bm{x}) \right) \right) \right),
\end{split}
\end{align}
where $J_1^{\prime}$ is the index set of the spectra of the Perron--Frobenius operator analogous to $J_1$, and $\hat{\bm{U}}^{\prime}$ and $\hat{\bm{V}}^{\prime}$ are the corresponding matrices. $\widehat{\bm{\psi}^{-1}}$ is another surrogate function for approximating the inverse function of $\bm{\psi}$. The new point $\bm{z}^{\star}$ reflects the convergent behavior of the dynamics over an invariant set. We refer to this type of models as \emph{two-tier} SIMs, which come from the structure involving two-tier maps by $\bm{\varphi}$ and $\bm{\psi}$. Two-tier SIMs are also represented as finite-depth models.

\vspace*{2mm}
Figure~\ref{illus:overview} shows the overview of SIMs. The black arrows denote the first transformation with the basis functions $\bm{\varphi}$. The blue and red arrows correspond to the transformations of the single-tier and two-tier SIMs, respectively. Each model considers the convergent behavior of the dynamics in the first-tier or second-tier space and then returns to the original space with the surrogate functions.

Although DEQs are defined by the transformation $\bm{f}_\theta$, SIMs require the specification of the basis functions $\bm{\varphi}$ and $\bm{\psi}$. One natural question may be how the two approaches and convergence properties relate to each other. Generally, if two dynamical systems converge to similar fixed points or invariant sets, the corresponding dynamics also have similar topological properties. However, the actual transformations are determined by many factors (e.g.\ the training algorithms and choice of basis functions), and the convergence regions of the two approaches will be different. Investigating how $\bm{f}_{\bm{\theta}}$ can be associated with $\bm{\varphi}$ and $\bm{\psi}$ is an important future work.

\subsection{Practical Implementation}
\label{ssec:imp}

The key to encode high expressiveness in the dynamics with SIMs is to utilize rich basis functions. Therefore, we leverage neural networks with structures to construct the components in our models~\eqref{eq:single-tier-model} and~\eqref{eq:two-tier-model} as follows.

First, we utilize a neural network whose inputs are only $\bm{x}$, which we denote by $\bm{\mu}_{\text{NN}}(\bm{x})$, to approximate $\bm{\varphi}(\bm{z}_0, \bm{x})$. This is because the initial state $\bm{z}_0$ can be basically assumed to be zero. Although we can prepare a non-zero $\bm{z}_0$ (e.g.\ the original implementation of DEQs allows to use the final state of the previous step in the training loop), the effectiveness of such a biased initial state is unclear. Therefore, we assume that $\bm{z}_0$ is always zero and encoded in the basis functions, and thus drop $\bm{z}_0$ from the input of $\bm{\varphi}$.

Next, because $\hat{\bm{U}}$ and $\hat{\bm{V}}$ in Eq.~\eqref{eq:single-tier-model} are simply real matrices, we can deal with them as two consecutive linear layers although $\hat{\bm{U}}$ and $\hat{\bm{V}}$ respectively consist of linearly independent vectors. Although we could consider the constraint, we ignore it because the vectors in a learned matrix will rarely be linearly dependent. $\hat{\bm{V}}$ has another implicit constraint: $\hat{\bm{V}}$ originates from $\bm{U}^{-1}$ and thus is affected by $\hat{\bm{U}}$. However, $\hat{\bm{V}}$ is also influenced by the vectors of $\bm{U}$ which correspond to eigenvalues with an absolute value less than 1. Although these vectors disappear by taking the limit and do not appear in the model, $\hat{\bm{V}}$ will have degrees of freedom thanks to the vectors. From this observation, we model $\hat{\bm{V}}$ independently from $\hat{\bm{U}}$.

Further, we prepare another neural network for the surrogate functions, $\widehat{\bm{\varphi}_{\bm{z}}^{-1}}$ and $\widehat{\bm{\psi}^{-1}}$. This network is, principally, required to approximate the inverse function corresponding to the respective part of $\bm{\varphi}$ and $\bm{\psi}$. However, even approximately modeling a neural network to be invertible is costly and difficult unless the dimension of the lifted space is identical to that of the original space~\citep{PNRM21}. The invertibility will also require SIMs to consider $\widehat{\bm{\varphi}_{\bm{x}}^{-1}}$ that is omitted from Eq.~\eqref{eq:deq-approx}. This is because it is necessary to construct $\widehat{\bm{\varphi}^{-1}}$ including $\widehat{\bm{\varphi}_{\bm{x}}^{-1}}$ to approximate the invertibility even though the output of $\widehat{\bm{\varphi}_{\bm{x}}^{-1}}$ is never used in subsequent transformations. In this paper, we focus on the practical aspect of our implementation and approximate $\widehat{\bm{\varphi}_{\bm{z}}^{-1}}$ in Eq.~\eqref{eq:single-tier-model} or the composition of $\widehat{\bm{\varphi}_{\bm{z}}^{-1}}$ and $\widehat{\bm{\psi}^{-1}}$ in Eq.~\eqref{eq:two-tier-model} by a neural network $\bm{\nu}_{\text{NN}}$ without the restriction of invertibility. This approach follows the same manner as common encoder-decoder models; an encoder model is usually designed independently of the corresponding decoder model despite their close connection of going back and forth between original and latent spaces.

For two-tier SIMs, we employ random Fourier features (RFFs)~\citep{RR07} to approximate the embedding $\bm{\psi}$, i.e.\
\begin{align*}
    \bm{\psi}(\bm{x})
    &=  \frac{1}{\sqrt{M/2}}
    \left(
          \sin(\bm{\omega}_1^{\top}\bm{x}),\
          \cos(\bm{\omega}_1^{\top}\bm{x}),\
          \cdots,\
          \sin(\bm{\omega}_{M/2}^{\top}\bm{x}),\
          \cos(\bm{\omega}_{M/2}^{\top}\bm{x})
        \right)^{\top}, \\
        \quad
    \bm{\omega}_j &\sim P(\bm{\omega}),
\end{align*}
where $\bm{\omega}_j \in \mathbb{R}^N$ for $j=1,\ldots,M/2$ are random vectors drawn from a probability distribution $P(\bm{\omega})$, to avoid the increase of the computational cost along the sample size when using reproducing kernels. Because $\bm{\psi}$ is the $M$-dimensional real-valued function, $\hat{\bm{U}}^{\prime}$ and $\hat{\bm{V}}^{\prime}$ are defined as $M \times K^{\prime}$ and $K^{\prime} \times M$ real matrices, respectively, where $K^{\prime}$ is the size of $J_1^{\prime}$. We model those matrices in the same way as $\hat{\bm{U}}$ and $\hat{\bm{V}}$.

Putting the above pieces together, we model Eqs.~\eqref{eq:single-tier-model} and~\eqref{eq:two-tier-model} by $\bm{z}_{\text{single-tier}}$ and $\bm{z}_{\text{two-tier}}$, respectively, as follows:
\begin{align}
\bm{z}_{\text{single-tier}} &= \bm{\nu}_{\text{NN}} \left( \hat{\bm{U}}\hat{\bm{V}} \bm{\mu}_{\text{NN}}(\bm{x}) \right) ~~\text{and}~~ \label{eq:single-tier-nn}\\
\bm{z}_{\text{two-tier}} &= \bm{\nu}_{\text{NN}} \left( \hat{\bm{U}}^{\prime}\hat{\bm{V}}^{\prime} \bm{\psi} \left(\bm{\mu}_{\text{NN}}(\bm{x}) \right) \right). \nonumber
\end{align}
In the end, those implementations of SIMs are realized as feedforward models. The implementations no longer require any root-finding algorithms for the forward pass and implicit differentiation for the backward pass. In the following experiments, we will instantiate more specific examples of the implementations for each task. However, the implementations do not limit themselves to such particular forms and have the flexibility to take various configurations by changing $\bm{\mu}_{\text{NN}}$ and $\bm{\nu}_{\text{NN}}$.

A practical merit of DEQs compared to feedforward models is the memory efficiency: the required memory does not depend on the number of transformations by $\bm{f}_{\bm{\theta}}$ owing to root-finding and implicit differentiation. Because we realize SIMs as feedforward models, they do not fully inherit this efficiency. However, DEQs also model $\bm{f}_{\bm{\theta}}$ with feedforward architectures such as Transformers~\citep{VSP+17,BKK19a} and thus need the memory for $\bm{f}_{\bm{\theta}}$, which could be potentially large. Additionally, we notice that DEQs are infinitely deep. The naive approximation with a feedforward model may need to increase the depth of the feedforward model. Our derivation showed that it is critical to approximate the DEQs according to the form of Eq.~\eqref{eq:single-tier-nn}. If the basis functions of SIMs could be represented by shallower models, it will save a lot of memory compared to naive approximations. We can regard SIMs as an intermediate approach that can approximate infinitely deep models while avoiding the simple dependence on depth.

\subsection{Relation to Implicit Neural Representation}
\label{ssec:inr}

Implicit neural representation has recently gained interest in the machine learning community. The aim is to represent complex natural signals (e.g.\ images, 3D objects, and audio waves) with a function modeled by a neural network~\citep{SMB+20,TSM+20}.

SIMs have an interesting connection to implicit neural representation by \citep{TSM+20}.
\citet{TSM+20} proposed to use a Fourier feature mapping as a pre-processing.
If the basis functions modeled by $\bm{\mu}_{\text{NN}}$ of single-tier SIMs are modeled by an RFF $\bm{\psi}$, then the implementation of the SIMs is the same as the Fourier feature mapping by \citep{TSM+20} up to the term $\hat{\bm{U}}\hat{\bm{V}}$:
\begin{align}
\label{eq:tier-rff}
\bm{z}_{\text{RFF}} &= \bm{\nu}_{\text{NN}} \left( \hat{\bm{U}}\hat{\bm{V}} \bm{\psi}(\bm{x}) \right).
\end{align}
As in Eq.~\eqref{eq:single-tier-nn}, $\bm{z}_0$ is omitted from $\bm{\psi}(\bm{x})$ because it is supposed to be zero and does not affect the output of $\bm{\psi}$. Alternatively, this RFF only model can be interpreted as an instance of two-tier SIMs when $\bm{\mu}_{\text{NN}}$ is an identity mapping. RFFs have been initially proposed as an approximation for kernel methods, which are also applied for basis functions in Koopman operator analysis~\citep{Kaw16}. Hence, RFFs will be a natural choice as a class of basis functions for SIMs. Eq.~\eqref{eq:tier-rff} indicates that such a reasonable choice leads to a similar method to \citep{TSM+20}.

\section{Related Works}
\label{sec:relwork}

The origin of DEQs dates back to the work on recurrent backpropagation (RBP)~\citep{Alm87, Pin87}. Those studies proposed to utilize early implicit-depth models and have been applied to other studies, e.g.\ graph neural networks~\citep{GMS05, SGT+08}. Recently, \citet{LXF+18} revisited the RBP algorithm and improved it with the conjugate gradient method and Neumann series. \citet{BKK19a} introduced a perspective of the use of a fixed point as a replacement for depth and called the approaches the DEQ. This study also proved the universality of a single DEQ layer and developed a practical quasi-Newton method that works for large-scale sequential tasks. Subsequent studies have reported the theoretical analysis~\citep{Kaw21, PWK21} and proposed improved architectures~\citep{BKK20, XWL+21}. Several relevant studies focused on the stability issue of DEQs. \citet{WK20} proposed monotone DEQs (monDEQs) with guaranteed convergence to a fixed point based on monotone operators. \citet{BKK21} also proposed to stabilize the training using Jacobian regularization.

The Koopman operator has been known for a long time as a tool for analyzing dynamics in physics~\citep{Koo31}. It has recently received attention that the spectral properties of the Koopman operator play an important role in revealing global characteristics of the underlying dynamics~\citep{Mez05}. Because the Koopman operator is linear but infinite-dimensional, a major challenge is how to compute the spectra of the Koopman operator in practice. Dynamic mode decomposition (DMD)~\citep{Sch10} has gained popularity as a data-driven approach to computing a reasonable finite-dimensional approximation of the Koopman spectra~\citep{RMBS+09}. While the original DMD algorithm supposes that the basis functions are linear, several subsequent studies have been proposed to utilize nonlinear basis functions~\citep{WKR15}, a kernel-based approach~\citep{Kaw16}, a Bayesian formulation~\citep{TKT+17}, and learning from data~\citep{TKY17}.

The Koopman operator is also applied to machine learning and the related fields. \citet{DR20} leveraged the representation with the Koopman operator to accelerate the training of neural networks. \citet{MFM+20} analyzed the dynamics in the training of neural networks with the Koopman operator and proposed a method to characterize the architectures of neural networks with the Koopman spectrum. \citet{TK21} proposed a neural network to learn stable dynamical systems by utilizing a map obtained by, for example, an eigenfunction of the Koopman operator (although it is not described explicitly in the paper) by extending stable deep dynamics models by \citet{MK19}.

\section{Experiments}
\label{sec:expr}

We evaluated SIMs through experiments on three supervised learning tasks. For all tasks, we first trained models with candidates of hyper-parameters on training data and evaluated them on validation data. We then selected the best hyper-parameter, re-trained the model with the best one, and evaluated the trained model on test data.
We implemented our proposed models and training algorithms using Pytorch~\citep{PGM+19}. For the basic building blocks and optimizers, we used the existing implementation in Pytorch. We tuned the hyper-parameters using Tune~\citep{LLN+18}. In all experiments, we set the search algorithm to the random search and trial scheduler to the asynchronous successive halving algorithm (ASHA)~\citep{LJR+20}. We conducted the experiments on an internal computing server with Intel Xeon Bronze 3104 CPUs and NVIDIA V100 GPUs.

We set $\bm{\mu}_{\text{NN}}$ as a task-specific neural network for each task. For $\bm{\psi}$, we used a normal distribution to sample random vectors. For $\bm{\nu}_{\text{NN}}$, we used a three-layer fully connected network (FCN) with ReLU activation~\citep{NH10} for all tasks in common. We used a linear function as $\bm{h}$ to fit the dimension of the hidden state to the output size and then the softmax function if the addressed task is a classification problem. For SIMs, we set $K=N/2$ and $K^{\prime}=M/2$, respectively.

The details of the datasets, model architectures, and training algorithms are described in~\ref{sec:app-ex}.

\subsection{Copy Memory Task}
\label{ssec:cpmem}

We first report the results of the copy memory task~\citep{HS97} to evaluate the effectiveness of SIMs against DEQs. The goal of the copy memory task is to predict a sequence of digits of length $T+20$ from another input sequence of the same length. The first ten elements of the input sequence consist of digits randomly drawn from 1 to 8, the subsequent $T-1$ elements are filled with 0, and the last eleven elements are all 9. Given this input, the first $T+10$ elements of the output sequence are 0 and the last ten elements are the same as the first ten ones of the input sequence. Hence, this task evaluates how well a model can remember the first elements of an input. In the experiment, we set $T$ to 500 and followed the experimental procedure of \citep{BKK18}.

For SIMs, we applied the temporal convolutional network (TCN) architecture to model $\bm{\mu}_{\text{NN}}$ by following the implementation of~\citep{BKK18}: the architecture includes 1D dilated causal convolution, ReLU activation, and residual connection~\citep{HZR+16}. Because the method of applying the TCN is rather complicated compared to other tasks, we explain the case of the single-tier SIM as an example. Formally, if we denote $\bm{x} \in \{0,1,\ldots,9\}^{520}$ as an input sequence, the TCN outputs the sequence of states:
\begin{align*}
\bm{W} = \text{TCN}(\bm{x}),
\end{align*}
where $\bm{W} = (\bm{w}_1,\ldots,\bm{w}_{520}) \in \mathbb{R}^{N \times 520}$ and $\bm{w}_j \in \mathbb{R}^{N}$ denotes a state on the lifted space at the $j$-th position in the sequence. Each state feeds the common architecture:
\begin{align*}
\bm{z}_{\text{single},j} &= \bm{\nu}_{\text{NN}} \left( \hat{\bm{U}}\hat{\bm{V}} \bm{w}_j \right)~~~(j=1,....,520),
\end{align*}
where $\bm{z}_{\text{single},j} \in \mathbb{R}^{d}$ is the hidden state of the $j$-th position. The prediction at the $j$-th position of the output sequence is obtained from $\bm{z}_{\text{single},j}$. We can interpret this architecture as follows; for this task, the model has different $\bm{\mu}_{\text{NN}}$ for each position, but it transforms all the positions to the hidden states at once by the TCN, and then the same subsequent transformation is applied to all the positions. Although we could use different $\bm{\nu}_{\text{NN}}$ and $\hat{\bm{U}}\hat{\bm{V}}$ for each position, the above architecture can retain memory in practice while keeping the model size small.

For the DEQ, we applied the Universal Transformer~\citep{DGV+19} as the base function $\bm{f}_\theta$ and mostly adopted the original implementation of~\citep{BKK19a}. However, the public source code was optimized for the tasks of language modeling; hence, we mainly modified the following two parts of the implementation. First, the implementation uses the adaptive softmax function~\citep{BA19,GJC+J17} to address the large word vocabulary. Because the copy memory task considers only ten digits, we did not use this architecture in the experiment. Second, the implementation uses the memory padding and nonzero initial hidden states that employ the final states of the previous step in a training loop. We tested the two techniques but observed that the test loss was considerably worse even though the training loss was fine. Hence, we omit the techniques and instead used the empty memory and zero initial hidden states in the experiment.

\begin{figure}[t]
    \centering
    \includegraphics{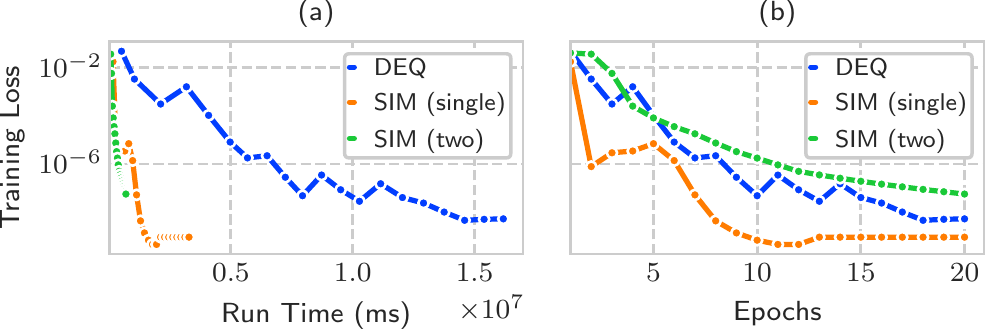} 
    \caption{Training losses along (a) run-time, and (b) epochs for three compared methods in the copy memory task.}
    \label{fig:cpmem}
\end{figure}

Table~\ref{ta:cpmem} shows the test cross-entropy loss per sequence and the number of learnable parameters for each model. Although the DEQ achieved the lowest loss, the single-tier SIM obtained was comparable with almost the same number of learnable parameters. Figure~\ref{fig:cpmem} (a) shows the progress of the training losses along the run-time for 20 epochs when we re-trained the models. The training speed depends on the batch size. During the hyper-parameter search, 1, 1, and 10 were selected as the batch size for the DEQ, single-tier SIM, and two-tier SIM, respectively. Hence, the two-tier SIM was the fastest to complete the training. For the DEQ and single-tier SIM, the single-tier model was ten times faster than the DEQ under the same batch size. Figure~\ref{fig:cpmem} (b) shows the progress of the training losses along the epochs. We can observe that the single-tier SIM converged faster than the DEQ.

\begin{table}[t]
    \caption{Test cross-entropy loss (test loss) and the number of learnable parameters (\#params) in the copy memory task.}
    \label{ta:cpmem}
    \centering
    \footnotesize
    \begin{tabular}{lrr}
        \toprule
                       & Test loss & \#params \\
        \midrule
          DEQ          & 2.24e-09  & 17,010 \\
          SIM (single) & 7.03e-09  & 17,294 \\
          SIM (two)    & 5.06e-08  & 17,294 \\
        \bottomrule
    \end{tabular}
\end{table}

\begin{table}[t]
    \caption{Mean of the test accuracy over three runs (test acc.) and the number of learnable parameters (\#params) in the image classification task. We showed the results of monDEQs reported in the paper~\citep{WK20}.}
    \label{ta:imgcl}
    \centering
    \scriptsize
    \begin{tabular}{lrrrrrr}
        \toprule
        \multicolumn{1}{c}{} & \multicolumn{2}{c}{CIFAR-10} & \multicolumn{2}{c}{SVHN} & \multicolumn{2}{c}{MNIST} \\
        \midrule
                          & Test acc.    & \#params &    Test acc. & \#params &    Test acc. & \#params \\ \midrule
        monDEQ (single)   & 74.0$\pm$0.1 &  172,218 & 88.7$\pm$1.1 &  172,218 & 99.1$\pm$0.1 &   84,460 \\
        monDEQ (multi)    & 72.0$\pm$0.3 &  170,194 & 92.4$\pm$0.1 &  170,194 & 99.0$\pm$0.1 &   81,394 \\
        SIM (single)      & 79.4$\pm$0.2 &  168,694 & 91.8$\pm$0.2 &  168,694 & 99.4$\pm$0.0 &   81,480 \\
        SIM (two)         & 78.2$\pm$0.2 &  168,264 & 92.4$\pm$0.1 &  168,264 & 99.4$\pm$0.0 &   80,466 \\
        \bottomrule
    \end{tabular}
\end{table}

\subsection{Image Classification}
\label{ssec:imgcl}

We next report the results of the image classification to compare SIMs to monDEQs. We followed the experiment of image classification conducted in~\citep{WK20}. We prepared the CIFAR-10~\citep{Kri09}, SVHN~\citep{NWC+11}, and MNIST~\citep{LBB+98} datasets, which contain images in 10 different classes and evaluated the classification performance in the standard setting. Following~\citep{WK20}, we evaluated the models on test data three times with different initialization and reported the averaged performance.

For SIMs, we employed convolutional neural networks for $\bm{\mu}_{\text{NN}}$. Following the VGG models~\citep{SZ15}, $\bm{\mu}_{\text{NN}}$ consists of two convolutional layers each of which has two convolution filters with ReLU activation and batch normalization and one max pooling. The output of the convolution layers is additionally transformed by a linear layer to fit the dimension of the lifted space. We constrained the number of learnable parameters of SIMs to be comparable to the one of monDEQs in~\citep{WK20}.

Table~\ref{ta:imgcl} lists the means of the test accuracy over three runs of monDEQs and SIMs with different initialization. SIMs showed comparable or better performances against monDEQs for all datasets. Particularly, the performance was improved for the CIFAR-10 dataset even though the SIMs have a similar number of learnable parameters to monDEQ ones.

\begin{table}[t]
    \caption{Mean of the test PSNR over 16 images of two types of datasets in the image regression task.}
    \label{ta:imgreg}
    \centering
    \footnotesize
    \begin{tabular}{lrr}
        \toprule
        \multicolumn{1}{c}{} &   \multicolumn{2}{c}{Test PSNR} \\ \midrule
                             &        Natural &           Text \\ \midrule
        SIM (single)         & 19.67$\pm$2.87 & 17.54$\pm$2.07 \\
        SIM (two)            & 22.24$\pm$3.03 & 23.30$\pm$2.72 \\
        SIM (RFF only)       & 25.19$\pm$3.92 & 27.69$\pm$1.63 \\
        \bottomrule
    \end{tabular}
\end{table}

\subsection{Image Regression}
\label{ssec:imgreg}

Finally, we report the results of the image regression task, which is an example of implicit neural representation tasks. In Section~\ref{ssec:inr}, we found that SIMs have a close connection to the work of~\citep{TSM+20}. The purpose of the experiment is to verify 1) that the RFF only model actually works well for this task, and 2) how well the other types of SIMs perform. The goal of the task is to obtain a neural network where the input is a 2-D pixel coordinate and the output is its 3-D RGB value. Following~\citep{TSM+20}, we evaluate SIMs with 32 datasets, where 16 are natural images~\footnote{\url{https://drive.google.com/uc?id=1TtwlEDArhOMoH18aUyjIMSZ3WODFmUab}} and the rest are text images~\footnote{\url{https://drive.google.com/uc?id=1V-RQJcMuk9GD4JCUn70o7nwQE0hEzHoT}}. This task considers each of the images to be one dataset. For an image, we picked 1/4 pixels as training data and other 1/4 as test data. Additionally, we prepared another 1/4 pixels as validation data to select the hyper-parameters.

We compare three instances of SIMs. The first two models employ an FCN for $\bm{\mu}_{\text{NN}}$: it consists of two linear layers with ReLU activation and one linear layer to fit the dimension of the lifted space. The last one is the RFF only model~\eqref{eq:tier-rff}.

\begin{figure}[t]
    \centering
    \begin{tabular}{ccc}
      \begin{minipage}[t]{0.28\hsize}
        \centering
        \includegraphics[scale=0.35]{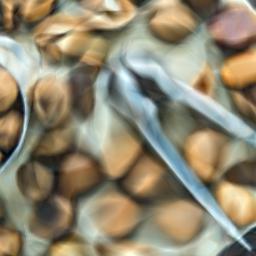}
      \end{minipage} &
      \begin{minipage}[t]{0.28\hsize}
        \centering
        \includegraphics[scale=0.35]{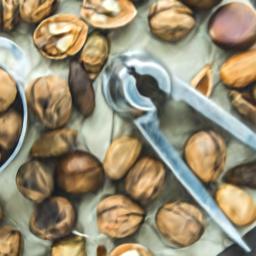}
      \end{minipage} &
      \begin{minipage}[t]{0.28\hsize}
        \centering
        \includegraphics[scale=0.35]{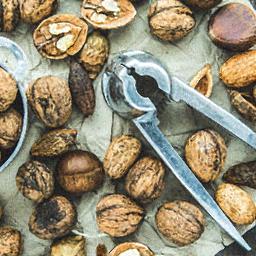}
      \end{minipage} \\
      \begin{minipage}[t]{0.28\hsize}
        \centering
        \includegraphics[scale=0.35]{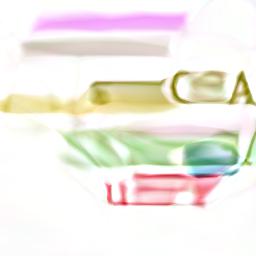}
        \caption*{\small{SIM (single)}}
      \end{minipage} &
      \begin{minipage}[t]{0.28\hsize}
        \centering
        \includegraphics[scale=0.35]{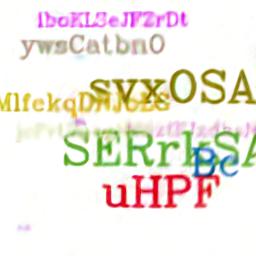}
        \caption*{\small{SIM (two)}}
      \end{minipage} &
      \begin{minipage}[t]{0.28\hsize}
        \centering
        \includegraphics[scale=0.35]{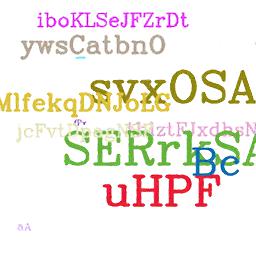}
        \caption*{\small{SIM (RFF only)}}
      \end{minipage}
    \end{tabular}
    \caption{Examples of prediction in the image regression task.}
    \label{fig:imgreg}
\end{figure}

Table~\ref{ta:imgreg} shows the mean of the test PSNR over 16 images of two types of datasets. The results indicate that the RFF only model was better than the single-tier SIM. This result is consistent with the original work on~\citep{TSM+20}. The two-tier SIM was also effective although the test PSNRs were slightly worse than the RFF only model. Figure~\ref{fig:imgreg} illustrates examples of prediction for two images. Although the single-tier SIM produced blurred images, the RFF only model can generate sharper ones. Although the images of the two-tier SIM are a little blurry, the detail can be recognized compared to the single-tier SIM.

%

\subsection{Discussion}

The first two results indicate that SIMs can provide time-effective alternatives to DEQs. This is because SIMs are implemented as feedforward models, which do not require implicit differentiation even though they can approximate DEQs. While we compared SIMs to DEQs as fairly as possible by keeping the number of parameters nearly the same, SIMs performed slightly better than DEQs in many experiments. It may be because the chosen basis functions were suitable for the tasks, and finite-depth models were easier to evaluate and optimize. The results also showed that two-tier SIMs did not necessarily outperform single-tier SIMs for the first two tasks. One of the reasons would be that even single-tier SIMs suffice for the tasks where the models can capture the essential feature of sequences and images.

The third result demonstrated that SIMs are also effective in the task of implicit neural representation. Particularly, we observed that the two-tier SIM also obtained moderate performance. The work of \citep{TSM+20} does not cover the form of the two-tier models, and we believe that this result is an interesting finding.

\section{Conclusions}
\label{sec:conc}

In this study, we considered DEQs from the viewpoint of the Koopman operator, which enables us to identify stable dynamics described by DEQs via the representation of the spectra. This perspective yielded our proposed SIMs that approximated DEQs and exploited more general dynamics converging to an invariant set. Despite having such noteworthy properties, the resulting models can be represented as simple feedforward models, which will provide new insights into the studies on DEQs.

A promising future work will be to consider more theoretical analysis. Investigating expressive powers, sample efficiency, and approximation error bounds would help to further understand SIMs. Another direction will be to explore ways to exploit dynamics converging to an invariant set. While we proposed the second-tier SIMs in this paper, other approaches will be possible and could improve the performance.

\section*{Acknowledgements}
\label{sec:ack}

This work was supported by JST CREST Grant Number JPMJCR1913 and JSPS KAKENHI Grant Numbers 18H03287, 22H00516, and 22K17950.

\appendix

\section{Convergent Behavior on Lifted Space}
\label{sec:app-deriv-limit}

We clarify the convergent behavior of the lifted dynamics in Section~3.
It should be noted that the following result is not novel; it employs a known consequence of linear algebra.

First, we define the $l$-th step of the lifted dynamics as
\begin{align*}
\bm{\varphi}^l \coloneqq \bm{A}^{l} \bm{\varphi}(\bm{z}_0,\bm{x}),
\end{align*}
where $\bm{\varphi}^0 \coloneqq \bm{\varphi}(\bm{z}_0,\bm{x})$.
To reveal the property of $\bm{\varphi}^l$, we decompose $\bm{A}^l$ with the eigenvalues and eigenvectors.

We begin with the decomposition of $\bm{A}$. If $\bm{A}$ is diagonalizable, it can be represented by the set of the eigenvalues and corresponding eigenvectors. However, if $\bm{A}$ has repeated eigenvalues, $\bm{A}$ is not necessarily diagonalizable. Although the Jordan canonical form can be used in such a case, this form is constructed by complex matrices if $\bm{A}$ contains complex eigenvalues. To represent $\bm{A}$ by real matrices for convenience, we consider the real Jordan canonical form: any real square matrix can be written as
\begin{align*}
\bm{A} = \bm{U} \bm{J} \bm{U}^{-1},
\end{align*}
where $\bm{U}$ is defined in Section 3.2, and $\bm{J} \in \mathbb{R}^{N \times N}$ is the following block diagonal matrix:
\begin{align*}
\bm{J} = \text{diag}(\bm{J}_1,\bm{J}_2,\ldots,\bm{J}_R),
\end{align*}
where $R \leq N$ corresponds to the number of linearly independent eigenvectors. Each block $\bm{J}_r \in \mathbb{R}^{s_r \times s_r}$ ($r=1,\ldots,R$) is associated with $s_r$ repeated eigenvalues with a value $\lambda_{(r)}$ if $\lambda_{(r)}$ is real or $s_r/2$ repeated eigenvalues with a value $\lambda_{(r)}$ and their $s_r/2$ complex conjugates with a value $\overline{\lambda_{(r)}}$ if $\lambda_{(r)}$ is nonreal. If $\lambda_{(r)}$ is real, $\bm{J}_r$ is also associated with one ordinal eigenvector of $\lambda_{(r)}$ and $s_r-1$ non-ordinal generalized-eigenvectors. If $\lambda_{(r)}$ is nonreal, $\bm{J}_r$ is associated with a complex conjugate pair of ordinal eigenvectors of $\lambda_{(r)}$ and $\overline{\lambda_{(r)}}$ and $(s_r-1)/2$ complex conjugate pairs of non-ordinal generalized-eigenvectors. It should also be noted that there always exists a conjugate eigenvalue for every nonreal eigenvalue when $\bm{A}$ is real. $\bm{J}_r$ takes one of the following two forms:
\begin{align*}
\bm{J}_r =
\begin{cases}
\bm{J}_{r, \text{R}} & \text{if $\lambda_{(r)}$ is real}\\
\bm{J}_{r, \text{C}} & \text{if $\lambda_{(r)}$ is nonreal},
\end{cases}
\end{align*}
where
\begin{align*}
\bm{J}_{r, \text{R}} =
\begin{pmatrix}
\lambda_{(r)} & 1 &  &  & \\
& \lambda_{(r)} & 1 & & \\
& & \ddots & \ddots & \\
& & & \lambda_{(r)} & 1 \\
& & & & \lambda_{(r)}
\end{pmatrix},
\end{align*}
\begin{align*}
\bm{J}_{r, \text{C}} =
\begin{pmatrix}
\bm{C}_{r} & \bm{I} &  &  & \\
& \bm{C}_{r} & \bm{I} & & \\
& & \ddots & \ddots & \\
& & & \bm{C}_{r} & \bm{I} \\
& & & & \bm{C}_{r}
\end{pmatrix}.
\end{align*}
Here, $\bm{C}_{r}$ and $\bm{I}$ are defined by
\begin{align*}
\bm{C}_{r} =
\begin{pmatrix}
\alpha_{(r)} & \beta_{(r)} \\
-\beta_{(r)} & \alpha_{(r)}
\end{pmatrix},~~~
\bm{I} =
\begin{pmatrix}
1 & 0 \\
0 & 1
\end{pmatrix},
\end{align*}
where
\begin{align*}
\lambda_{(r)}=\alpha_{(r)}+i\beta_{(r)} \\
\overline{\lambda_{(r)}}=\alpha_{(r)}-i\beta_{(r)}.
\end{align*}
As a special case, $\bm{J}_{r, \text{R}}$ can be a scalar $\lambda_{(r)}$ when $s_r=1$, and $\bm{J}_{r, \text{C}}$ can be $\bm{C}_{r}$ when $s_r=2$. In this case, $\bm{J}_{r, \text{R}}$ and $\bm{J}_{r, \text{C}}$ are only associated with one ordinal eigenvector and a complex conjugate pair of ordinal eigenvectors, respectively. Moreover, the matrix $\bm{C}_{r}$ can be rewritten as a polar form:
\begin{align*}
\bm{C}_{r} = r_{(r)}
\begin{pmatrix}
\cos\Delta_{(r)} & \sin\Delta_{(r)} \\
-\sin\Delta_{(r)} & \cos\Delta_{(r)}
\end{pmatrix},
\end{align*}
where
\begin{align*}
r_{(r)} &= |\lambda_{(r)}| = \sqrt{\alpha_{(r)}^2 + \beta_{(r)}^2}, \\
\Delta_{(r)} &= \arctan(\beta_{(r)}/\alpha_{(r)}).
\end{align*}

The real Jordan canonical form can represent $\bm{\varphi}^l$ as
\begin{align*}
\bm{\varphi}^l = \bm{U} \bm{J}^l \bm{U}^{-1} \bm{\varphi}(\bm{z}_0,\bm{x}),
\end{align*}
where
\begin{align*}
\bm{J}^l = \text{diag}(\bm{J}^l_1,\bm{J}^l_2,\ldots,\bm{J}^l_R).
\end{align*}
Each $\bm{J}_{r}^{l}$ can be written as
\begin{align}
\label{eq:real-block}
\bm{J}_{r, \text{R}}^{l} =
\begin{pmatrix}
\lambda_{(r)}^l & \binom{l}{1}\lambda_{(r)}^{l-1} & & \cdots & \binom{l}{s_r-1}\lambda_{(r)}^{l-s_r+1} \\
& \lambda_{(r)}^l & & \cdots & \binom{l}{s_r-2}\lambda_{(r)}^{l-s_r+2} \\
& & \ddots & \ddots & \vdots\\
& & & \lambda_{(r)}^l & \binom{l}{1}\lambda_{(r)}^{l-1} \\
& & & & \lambda_{(r)}^l
\end{pmatrix},
\end{align}
or
\begin{align}
\label{eq:nonreal-block}
\bm{J}_{r, \text{C}}^{l} = 
\begin{pmatrix}
\bm{C}_{r}^l & \binom{l}{1}\bm{C}_{r}^{l-1} &  & \cdots & \binom{l}{s_r/2-1}\bm{C}_{r}^{l-s_r/2+1} \\
& \bm{C}_{r}^l & \binom{l}{1}\bm{C}_{r}^{l-1} & \cdots & \binom{l}{s_r/2-2}\bm{C}_{r}^{l-s_r/2+2} \\
& & \ddots & \ddots & \vdots\\
& & & \bm{C}_{r}^l & \binom{l}{1}\bm{C}_{r}^{l-1} \\
& & & & \bm{C}_{r}^l
\end{pmatrix},
\end{align}
where
\begin{align*}
\bm{C}_{r}^l &= \left( r_{(r)}
\begin{pmatrix}
\cos\Delta_{(r)} & \sin\Delta_{(r)} \\
-\sin\Delta_{(r)} & \cos\Delta_{(r)}
\end{pmatrix} \right)^l \\
&= r_{(r)}^l
\begin{pmatrix}
\cos l\Delta_{(r)} & \sin l\Delta_{(r)} \\
-\sin l\Delta_{(r)} & \cos l\Delta_{(r)}
\end{pmatrix}.
\end{align*}
This form makes it easy to evaluate the convergent behavior of the lifted dynamics: because the number of steps $l$ only depends on the block diagonal matrix $\bm{J}^l$, it suffices to focus on each block of $\bm{J}^l$ that takes the form of Eq.~\eqref{eq:real-block} or~\eqref{eq:nonreal-block}.

If the spectral radius $\rho(\bm{A}) < 1$, all the elements in every block of $\bm{J}^l$ converge to 0 by taking the limit $l\to\infty$. Hence, the lifted dynamics in case~\ref{enu:case-origin} converges to the origin:
\begin{align*}
\lim_{l \to \infty} \bm{\varphi}^l
&= \bm{U} \bm{O}_N \bm{U}^{-1} \bm{\varphi}(\bm{z}_0,\bm{x}) \\
&= \bm{0},
\end{align*}
where $\bm{O}_N \in \mathbb{R}^{N \times N}$ is a zero square matrix of order $N$.

If $\rho(\bm{A}) = 1$, we can classify the limit of the $r$-th block $\bm{J}_r^l$ into the following six cases:
\renewcommand{\theenumi}{$\langle$\arabic{enumi}$\rangle$}
\begin{enumerate}
\item If $|\lambda_{(r)}| < 1$, then $\bm{J}_r^l$ converges to a zero matrix:
\begin{align*}
    \lim_{l \to \infty} \bm{J}_r^l = \bm{O}_{s_r}.
\end{align*}\label{enu:block-zero}
\item If $\lambda_{(r)} = 1$ and $s_r = 1$, then $\bm{J}_r^l$ converges to 1:
\begin{align*}
    \lim_{l \to \infty} \bm{J}_r^l = \lim_{l \to \infty} \lambda_{(r)}^l=1.
\end{align*}\label{enu:block-one}
\item If $\lambda_{(r)} = -1$ and $s_r = 1$, then $\bm{J}_r^l$ neither converges nor diverges and takes the form of
\begin{align*}
    \bm{J}_r^l=\lambda_{(r)}^l=(-1)^l.
\end{align*}\label{enu:block-mone}
\item If $\lambda_{(r)} \in \{1, -1\}$, and $s_r > 1$, then the non-diagonal elements of $\bm{J}_r^l (=\bm{J}_{r,R}^l)$, e.g.\ $\binom{l}{1}\lambda_{(r)}^{l-1}$, diverge.\label{enu:block-real-unstable}
\item If $|\lambda_{(r)}| = 1$, $\lambda_{(r)}$ is nonreal, and $s_r = 2$, then $\bm{J}_r^l$ neither converges nor diverges and takes the form of
\begin{align}
    \label{eq:rotate}
    \bm{J}_r^l = \bm{C}_r^l =
    \begin{pmatrix}
        \cos l\Delta_{(r)} & \sin l\Delta_{(r)} \\
        -\sin l\Delta_{(r)} & \cos l\Delta_{(r)}
    \end{pmatrix}.
\end{align}\label{enu:block-rotate}
\item If $|\lambda_{(r)}| = 1$, $\lambda_{(r)}$ is nonreal, and $s_r > 2$, then the non-block-diagonal parts of $\bm{J}_r^l (=\bm{J}_{r,C}^l)$, e.g.\ $\binom{l}{1}\bm{C}_{r}^{l-1}$, diverge.\label{enu:block-nonreal-unstable}
\end{enumerate}
Hence, if at least one block falls into cases~\ref{enu:block-real-unstable} or~\ref{enu:block-nonreal-unstable}, the lifted dynamics diverges: the complexity of the non-diagonal elements or non-block-diagonal parts is at least $\mathcal{O}(l)$. This case falls into case~\ref{enu:case-unstable}.

Case~\ref{enu:case-point} means all the blocks are either case~\ref{enu:block-zero} or~\ref{enu:block-one}, and each block in case~\ref{enu:block-one} corresponds to one of the eigenvalues in $J_1$. Hence, the eigenvectors with eigenvalues of $\lambda_j=1$ are linearly independent. In the end, $\bm{J}^l$ converges to a diagonal matrix $\bm{P}_{J_1} \in \mathbb{R}^{N \times N}$ where the diagonal elements corresponding to $J_1$ are 1 and the rest are 0, and we obtain Eq.~(6):
\begin{align*}
\lim_{l \to \infty} \bm{\varphi}^l
&= \bm{U} \bm{P}_{J_1} \bm{U}^{-1} \bm{\varphi}(\bm{z}_0,\bm{x}) \\
&= \sum_{j \in J_1} \bm{u}_j \bm{v}_j^{\top} \bm{\varphi}(\bm{z}_0,\bm{x}).
\end{align*}

Case~\ref{enu:case-set} means that all the blocks are either case~\ref{enu:block-zero},~\ref{enu:block-one},~\ref{enu:block-mone}, or~\ref{enu:block-rotate}. Each block in cases~\ref{enu:block-mone} and~\ref{enu:block-rotate} corresponds to one of the eigenvalues of $J_2$ and the complex conjugate pairs of the eigenvalues of $J_3$, respectively. Hence, the eigenvectors with eigenvalues of $|\lambda_j|=1$ are linearly independent. In addition to $\bm{P}_{J_1}$, if we denote $\bm{P}_{J_2} \in \mathbb{R}^{N \times N}$ as a diagonal matrix where the diagonal elements corresponding to $J_2$ are -1 and the rest are 0 and $\bm{P}_{J_3} \in \mathbb{R}^{N \times N}$ as a block diagonal matrix where the diagonal blocks corresponding to the pairs of $J_3$ take the form of Eq.~\eqref{eq:rotate} and the rest are zero, we obtain Eq.~(7):
\begin{align*}
\lim_{l \to \infty} \bm{\varphi}^l
&= \lim_{l\to\infty} \bm{U} (\bm{P}_{J_1} + \bm{P}_{J_2} + \bm{P}_{J_3}) \bm{U}^{-1} \bm{\varphi}(\bm{z}_0,\bm{x}) \\
&= \lim_{l\to\infty} \Biggl( \sum_{j \in J_1} \bm{u}_j \bm{v}_j^{\top}
    + \sum_{j \in J_2} (-1)^l \bm{u}_j \bm{v}_j^{\top} \nonumber\\
    &  + \sum_{(j,k) \in J_3} \big(\cos(l\Delta_j)\bm{u}_{j}-\sin(l\Delta_k)\bm{u}_{k}\big) \bm{v}_{j}^{\top}
    \nonumber\\
    & + \big(\sin(l\Delta_j)\bm{u}_{j}+\cos(l\Delta_k)\bm{u}_{k}\big) \bm{v}_{k}^{\top} \Biggr)
    \bm{\varphi}(\bm{z}_0,\bm{x}).
\end{align*}

Lastly, if $\rho(\bm{A}) > 1$, at least one block of $\bm{J}^l$ diverges. This case falls into case~\ref{enu:case-unstable}.

\section{Experimental Details}
\label{sec:app-ex}

We show the detailed settings and configurations of the experiments in Section~\ref{sec:expr}. It should be noted that we used the default values of Pytorch for the arguments not mentioned below.

Table~\ref{ta:asha} shows that the settings of Tune and ASHA for each task. \#trials denotes the number of trials in running a search algorithm. Scope denotes the method of selecting the best hyper-parameters. We used \texttt{last} that selects the best one by comparing the last performance at the end of training. \#max\_t denotes the number of maximum epochs or iterations in a trial. ASHA periodically stops trials when the specified metric is poor and reduces them by a factor of the reduction factor. However, any trial is run until the grace period. Those settings are partially different for each task but the same as compared methods in a task.

Table~\ref{ta:psi} shows the detailed architecture of the reverse model $\bm{\nu}_{\text{NN}}$ in SIMs. The first column denotes the layers of $\bm{\nu}_{\text{NN}}$, which transform the input in order from the top to the bottom layers. ``$\times$ 2'' means that the same transformation has been repeated twice. The second column denotes detailed information about the layers. The variable \#hidden units is set to a different value for each task.

\begin{table}[t]
    \caption{Settings of Tune and ASHA for each task.}
    \label{ta:asha}
    \centering
    \scriptsize
    \begin{tabular}{lrrr}
        \toprule
                                &   Copy memory & Image classification & Image regression \\ \midrule
        \#trials                &           100 &                  200 &              200 \\
        Scope                   & \texttt{last} &        \texttt{last} &    \texttt{last} \\
        \#max\_t (ASHA)         &             6 &                   80 &              100 \\
        Metric (ASHA)           &          loss &             accuracy &             PSNR \\
        Grace period (ASHA)     &             3 &                    5 &                5 \\
        Reduction factor (ASHA) &             2 &                    2 &                2 \\
        \bottomrule
  \end{tabular}
\end{table}

\begin{table}[t]
    \caption{Architecture of the reverse model $\bm{\nu}_{\text{NN}}$ in SIMs for all tasks.}
    \label{ta:psi}
    \centering
    \scriptsize
    \begin{tabular}{lcc}
        \toprule
                                                & &                          Detail \\ \midrule
        \textbf{Fully connected layer} $\times$ 2 & &                                 \\
        \quad Linear                              & & output features: \#hidden units \\
        \quad ReLU                                & &                                 \\
        \textbf{Linear layer}                     & &                                 \\
        \quad Linear                              & & output features: \#hidden units \\
        \bottomrule
  \end{tabular}
\end{table}

\subsection{Copy Memory Task}
\label{ssec:app-cpmem}

Table~\ref{ta:dataset-cpmem} lists the statistics of the dataset in the copy memory task. \#train, \#valid, and \#test denote the number of training, validation, and test data, respectively. We first trained the models with the candidates of hyper-parameters on the training data and selected the best hyper-parameter on the validation data. We then used both training and validation data to re-train the model with the best hyper-parameter and evaluated it on the test data. A data point of this dataset consists of the pair of an input sequence of length 520 and an output sequence of length 520. We generated all the data points by following the procedure of~\citep{BKK18}.

Table~\ref{ta:varphi-cpmem} shows the model architecture of the TCN. The first building block is the temporal convolution layer where we adopted the architecture of~\citep{BKK18}: it mainly consists of the 1D dilated causal convolution, ReLU activation, and residual connection, and down-sampling with 1D convolution is applied to the residual connection~\citep{HZR+16} except for the first layer. We stacked this layer eight times and then connected a linear layer to fit the dimension of the lifted space.

Table~\ref{ta:config-cpmem} lists the configurations of the compared models. The upper and lower rows denote the configuration of the model architectures and training algorithms, respectively. The configuration with the set notation (e.g.\ [1e-3, 1e+3) and \{1, 2, 5, 10, 20\}) means that the corresponding variable was tuned as a hyper-parameter within the range of the set. For each trial, ASHA sampled a candidate from the set uniformly at random. For the configuration with a real interval, it was sampled in the logarithmic space with base 10. In this task, we tuned the bandwidth of the RFF, batch size, and learning rate. The configuration of the DEQ indicates that the variables of the implementation of~\citep{BKK19a}: we set it to ensure almost the same number of learnable parameters as SIMs. For the training algorithm of the DEQ, we adopted Adam~\citep{KB14} with the step-wise cosine decaying schedule for the learning rate as in~\citep{BKK19a}. For SIMs, we used Adam with a constant learning rate schedule and slightly larger $\epsilon$ to improve the stability of the training algorithm. We also used gradient clipping with the default values of the DEQ and TCN. For all compared models, we minimized the cross-entropy loss, i.e.\ the negative log-likelihood with the softmax function. Although every model contains the Dropout layer~\citep{SHK+14}, we did not apply it for all models.

\begin{table}[t]
    \caption{Statistics of the dataset in the copy memory task.}
    \label{ta:dataset-cpmem}
    \centering
    \scriptsize
    \begin{tabular}{rrrrr}
        \toprule
        \#train & \#valid & \#test &     Input dim. &    Output dim. \\ \midrule
          4,500 &     500 &    500 & 520 $\times$ 1 & 520 $\times$ 1 \\
        \bottomrule
  \end{tabular}
\end{table}

\begin{table}[t]
    \caption{Architecture of the TCN in SIMs for the copy memory task.}
    \label{ta:varphi-cpmem}
    \centering
    \scriptsize
    \begin{tabular}{lcc}
        \toprule
                                                       & & Details \\ \midrule
        \textbf{Temporal convolution layer} $\times$ 8 & &                                                \\
        \quad 1D dilated causal convolution            & & kernel size: 8, output channels: 10 \\
        \quad Weight normalization                     & &                                                \\
        \quad Chomp                                    & &                                                \\
        \quad ReLU                                     & &                                                \\
        \quad Dropout                                  & &                                                \\
        \quad 1D dilated causal convolution            & & kernel size: 8, output channels: 10 \\
        \quad Weight normalization                     & &                                                \\
        \quad Chomp                                    & &                                                \\
        \quad ReLU                                     & &                                                \\
        \quad Dropout                                  & &                                                \\
        \quad Residual connection                      & &                                                \\
        \quad ReLU                                     & &                                                \\
        \textbf{Linear layer}                          & &                                                \\
        \quad Linear                                   & & out features: $N$                              \\
        \bottomrule
  \end{tabular}
\end{table}

\begin{table}[t]
    \caption{Configurations of compared models in the copy memory task.}
    \label{ta:config-cpmem}
    \centering
    \scriptsize
    \begin{tabular}{lccc}
        \toprule
                                                     &                     DEQ &      SIM (single-tier) &         SIM (two-tier) \\ \midrule
        \textbf{Model architectures}   & & & \\
        \quad n\_head                                  &                       8 &                      - &                      - \\
        \quad d\_head                                  &                       5 &                      - &                      - \\
        \quad d\_model                                 &                      40 &                      - &                      - \\
        \quad d\_inner                                 &                      40 &                      - &                      - \\
        \quad pre\_lnorm                               &                    True &                      - &                      - \\
        \quad wnorm                                    &                    True &                      - &                      - \\
        \quad f\_thres                                 &                      30 &                      - &                      - \\
        \quad \#pretraining steps                      &                       0 &                      - &                      - \\
        \quad $N$                                      &                       - &                     32 &                     32 \\
        \quad $M$                                      &                       - &                      - &                     32 \\
        \quad Bandwidth of RFF                         &                       - &                      - &           [1e-3, 1e+3) \\
        \quad \#hidden units of $\bm{\nu}_{\text{NN}}$ &                       - &                     32 &                     32 \\
        \quad Dropout rate                             &                     0.0 &                    0.0 &                    0.0 \\
        \textbf{Training algorithms}                   &                         &                        &                        \\
        \quad Batch size                               &     \{1, 2, 5, 10, 20\} &    \{1, 2, 5, 10, 20\} &    \{1, 2, 5, 10, 20\} \\
        \quad \#epochs                                 &                      20 &                     20 &                     20 \\
        \quad Optimizer                                &                    Adam & Adam ($\epsilon$=1e-5) & Adam ($\epsilon$=1e-5) \\
        \quad Learning rate                            &            [1e-4, 0.05) &           [1e-4, 0.05) &           [1e-4, 0.05) \\
        \quad Learning rate schedule                   &      cosine (step-wise) &              constant  &               constant \\
        \quad Gradient clipping value                  &                    0.25 &                   1.0  &                    1.0 \\
        \bottomrule
  \end{tabular}
\end{table}

\subsection{Image Classification}
\label{ssec:app-imgcl}

Table~\ref{ta:dataset-imgcl} shows the statistics of three datasets in the image classification task. We pre-processed these datasets in almost the same way as~\citep{WK20}. The difference from~\citep{WK20} is to additionally prepare the validation data. We initially used 90$\%$ of the prepared training data as training data and the remaining 10$\%$ as validation data. We then used both of them to train models with the best hyper-parameters for evaluating the performance on the test data.

Table~\ref{ta:varphi-imgcl} shows the architecture of $\bm{\mu}_{\text{NN}}$ in SIMs. We first used a combination of convolution and max-pooling layers, which follows the VGG architecture~\citep{SZ15} and then applied a linear layer to fit the dimension of the lifted space.

Table~\ref{ta:config-imgcl} lists the configuration of SIMs. In this task, we tuned the bandwidth of the RFF, the batch size, the number of epochs, the learning rate, and the learning rate schedule. We set the model architectures of SIMs as almost the same size as that of the compared monDEQs. We also applied the training algorithm of the monDEQs to SIMs: the learning rate schedules are also implemented in the same way as the monDEQs. As in the monDEQs, we minimized the cross-entropy loss for SIMs.

\begin{table}[t]
    \caption{Statistics of three datasets in the image classification task.}
    \label{ta:dataset-imgcl}
    \centering
    \scriptsize
    \begin{tabular}{lrrrrr}
        \toprule
                 & \#train & \#valid & \#test & Input dim. & Output dim. \\ \midrule
        CIFAR-10 & 45,000 & 5,000 & 10,000 & 32 $\times$ 32 $\times$ 3 & 10 $\times$ 1 \\
        SVHN     & 65,931 & 7,326 & 26,032 & 32 $\times$ 32 $\times$ 3 & 10 $\times$ 1 \\
        MNIST    & 54,000 & 6,000 & 10,000 & 28 $\times$ 28 $\times$ 1 & 10 $\times$ 1 \\
        \bottomrule
  \end{tabular}
\end{table}

\begin{table}[t]
    \caption{Architecture of $\bm{\mu}_{\text{NN}}$ in SIMs for the image classification task.}
    \label{ta:varphi-imgcl}
    \centering
    \scriptsize
    \begin{tabular}{lcc}
        \toprule
                                              & & Details \\ \midrule
        \textbf{Convolution layer} $\times$ 2 & & \\
        \quad 2D convolution                  & & kernel size: 3, padding: 1, output channels: \#channels \\
        \quad Batch norm.                     & & \\
        \quad ReLU                            & & \\
        \textbf{Max pooling layer}            & & kernel size: 2, stride: 2 \\
        \textbf{Convolution layer} $\times$ 2 & & \\
        \quad 2D convolution                  & & kernel size: 3, padding: 1, output channels: \#channels \\
        \quad Batch norm.                     & & \\
        \quad ReLU                            & & \\
        \textbf{Max pooling layer}            & & kernel size: 2, stride: 2 \\
        \textbf{Linear layer}                 & & \\
        \quad Flatten                         & & \\
        \quad Linear                          & & out features: $N$ \\
        \bottomrule
  \end{tabular}
\end{table}

\begin{table}[t]
    \caption{Configurations of SIMs in the image classification task.}
    \label{ta:config-imgcl}
    \centering
    \scriptsize
    \begin{tabular}{lcc}
        \toprule
                                                     &                SIM (single-tier) &                   SIM (two-tier) \\ \midrule
        \textbf{Model architectures} & & \\
        \quad \#channels of $\bm{\mu}_{\text{NN}}$     &        24 (MNIST) or 38 (others) &        21 (MNIST) or 36 (others) \\
        \quad $N$                                      &                               50 &                               50 \\
        \quad $M$                                      &                                - &                              104 \\
        \quad Bandwidth of RFF                         &                                - &                     [1e-3, 1e+3) \\
        \quad \#hidden units of $\bm{\nu}_{\text{NN}}$ &                               32 &                               32 \\
        \textbf{Training algorithms} & & \\
        \quad Batch size                               &            \{64, 128, 256, 512\} &            \{64, 128, 256, 512\} \\
        \quad \#epochs                                 &               \{20, 40, 60, 80\} &               \{20, 40, 60, 80\} \\
        \quad Optimizer                                &                             Adam &                             Adam \\
        \quad Learning rate                            &                     [1e-4, 0.05) &                     [1e-4, 0.05) \\
        \quad Learning rate schedule                   &       \{1cycle, step, constant\} &       \{1cycle, step, constant\} \\
        \quad Step size for step schedule              & $\{5i \mid i \in (1,2,...10) \}$ & $\{5i \mid i \in (1,2,...10) \}$ \\
        \bottomrule
  \end{tabular}
\end{table}

\subsection{Image Regression}
\label{ssec:app-imgreg}

Table~\ref{ta:dataset-imgreg} shows the statistics of a dataset in the image regression task. All the images have 512 $\times$ 512 pixels, and we used equally spaced 1/4 (65,536) pixels from an image as training data, another 1/4 pixels as validation data, and other 1/4 pixels as test data. It should be noted that all the images are the same size and thus have the same statistics.

Table~\ref{ta:varphi-imgreg} shows the architecture of $\bm{\mu}_{\text{NN}}$ in SIMs. We used a neural network that consists of two fully connected layers and one additional linear layer to fit the dimension of the lifted space.

Table~\ref{ta:config-imgreg} shows the configuration of SIMs. In this task, we tuned the number of hidden units of $\bm{\mu}_{\text{NN}}$, the first and second lift dimensions (i.e.\ $N$ and $M$), the bandwidth of the RFF, and the learning rate. Following~\citep{TSM+20}, we applied Adam with the constant learning rate and performed the full batch gradient descent which uses all data points (pixels) for the update of parameters. \#iterations denotes the number of steps in the gradient descent and was set to 2000 as in~\citep{TSM+20}. We minimized the mean squared loss for all models. The difference from~\citep{TSM+20} is that we did not apply the sigmoid function before the output for each model: we observed such a setting improved the performance.

\begin{table}[t]
    \caption{Statistics of each dataset in the image regression task. It should be noted that all 32 datasets have the same statistics.}
    \label{ta:dataset-imgreg}
    \centering
    \scriptsize
    \begin{tabular}{rrrrr}
        \toprule
        \#train & \#valid & \#test & Input dim. & Output dim. \\ \midrule
         65,536 & 65,536 & 65,536 & 2 $\times$ 1 & 3 $\times$ 1 \\
        \bottomrule
  \end{tabular}
\end{table}

\begin{table}[t]
    \caption{Architecture of $\bm{\mu}_{\text{NN}}$ in SIMs for the image regression task.}
    \label{ta:varphi-imgreg}
    \centering
    \scriptsize
    \begin{tabular}{lcc}
        \toprule
                                                  & &                          Details \\ \midrule
        \textbf{Fully connected layer} $\times$ 2 & &                                  \\
        \quad Linear                              & & output features: \#hidden units \\
        \quad ReLU                                & &                                  \\
        \textbf{Linear layer}                     & &                                  \\
        \quad Linear                              & & output features: $N$             \\
        \bottomrule
  \end{tabular}
\end{table}

\begin{table}[t]
    \caption{Configurations of SIMs in the image regression task.}
    \label{ta:config-imgreg}
    \centering
    \tiny
    \begin{tabular}{lccc}
        \toprule
                                                       &         SIM (single-tier) &            SIM (two-tier) &            SIM (RFF only) \\ \midrule
        \textbf{Model architectures} & & \\
        \quad \#hidden units of $\bm{\mu}_{\text{NN}}$ & \{32, 64, 128, 256, 512\} & \{32, 64, 128, 256, 512\} &                         - \\
        \quad $N$                                      & \{32, 64, 128, 256, 512\} & \{32, 64, 128, 256, 512\} & \{32, 64, 128, 256, 512\} \\
        \quad $M$                                      &                         - & \{32, 64, 128, 256, 512\} &                         - \\
        \quad Bandwidth of RFF                         &                         - &              [1e-3, 1e+3) &              [1e-3, 1e+3) \\
        \quad \#hidden units of $\bm{\nu}_{\text{NN}}$ &                       256 &                       256 &                       256 \\
        \textbf{Training algorithms} & & \\
        \quad \#iterations                             &                      2000 &                      2000 &                      2000 \\
        \quad Optimizer                                &                      Adam &                      Adam &                      Adam \\
        \quad Learning rate                            &              [1e-4, 0.05) &              [1e-4, 0.05) &              [1e-4, 0.05) \\
        \quad Learning rate schedule                   &                  constant &                  constant &                 constant  \\
        \bottomrule
  \end{tabular}
\end{table}

\bibliographystyle{model5-names}\biboptions{authoryear}
\bibliography{references}

\end{document}